\documentclass[runningheads]{llncs}

\usepackage[T1]{fontenc}

\usepackage{graphicx}
\usepackage{CJK}
\usepackage{graphicx}
\usepackage{graphics}
\usepackage{epstopdf}
\usepackage{times}
\usepackage{latexsym}
\usepackage{stfloats}
\usepackage{amsfonts,amssymb}
\usepackage{bm}
\usepackage{multicol}
\usepackage{multirow}
\usepackage{booktabs}
\usepackage{arydshln}
\usepackage{amsmath}
\usepackage{hyperref}
\usepackage{algorithm}  
\usepackage{algorithmic}
\usepackage{ulem}
\usepackage{color}

\begin{document}

\title{An Adversarial Multi-Task Learning Method for Chinese Text Correction with Semantic Detection}

\titlerunning{Chinese Text Correction with Adversarial Multi-Task Learning}

\author{Fanyu Wang\orcidID{0000-0002-9937-8534} \and
Zhenping Xie*\orcidID{0000-0002-9481-9599}}
\authorrunning{Fanyu and Zhenping.}

\institute{School of Artificial Intelligence and Computer Science, Jiangnan University, Wuxi 214122, China \\ 
\email{xiezp@jiangnan.edu.cn}}
\maketitle 
\begin{abstract}
Text correction, especially the semantic correction of more widely used scenes, is strongly required to improve, for the fluency and writing efficiency of the text. An adversarial multi-task learning method is proposed to enhance the modeling and detection ability of character polysemy in Chinese sentence context. Wherein, two models, the masked language model and scoring language model, are introduced as a pair of not only coupled but also adversarial learning tasks. Moreover, the Monte Carlo tree search strategy and a policy network are introduced to accomplish the efficient Chinese text correction task with semantic detection. The experiments are executed on three datasets and five comparable methods, and the experimental results show that our method can obtain good performance in Chinese text correction task for better semantic rationality.

\end{abstract}
\section{Introduction}
\subsection{Bottlenecks and Defects}
Text correction is an essential process for daily writing, also, has been widely developed in current office software products~\cite{ghufron2018role,napoles-etal-2017-jfleg,omelianchuk-etal-2020-gector}. However, the complexity and flexibility of natural language especially for Chinese, cause huge obstacles to developing a high-quality text correction method. For Chinese, its finest process level of parsing sentences is character-level containing over 20,000 commonly used vocabulary characters. Usually, each sentence is composed of several phrases, and there is no characteristic separator between phrases, in which each phrase also contains several characters. Hence, the syntax in Chinese is complicated and easy to be misunderstood.

Traditionally, the two aspects in Chinese error correction are divided into CSC (Chinese Spelling Check) and CGEC (Chinese Grammatical Error Correction) which accord to their processing level as character-level and phrase-level respectively. The complexity syntax determines that there are few robust logic and adequate features for checking the grammar errors at phrase-level. At present, the pertained models like BERT (Bidirectional Encoder Representations from Transformers~\cite{devlin-etal-2019-bert}) is considered to represent the sentence semantics, and CRF (Conditional Random Field~\cite{Lafferty-JohnD}) can be introduced to realize the effective grammatical error correction~\cite{luo-etal-2020-chinese,li-shi-2021-tail}. For the character-level processing, the visual and phonological similarities is employed to generate the confusion set for Chinese Spelling Check~\cite{Liu-visually,wang-etal-2018-hybrid}. Moreover, the additional feature is also integrated into BERT to enhance the consistency of the Spelling Check methods~\cite{hong-etal-2019-faspell,cheng-etal-2020-spellgcn,Tan-Min-and-Chen-Dagang,ji-etal-2021-spellbert}. It reflects that the semantic feature is unreliable for Chinese Spelling Check, and the auxiliary verification methods utilized the the additional features is reasonably introduced.

\begin{table}[t]
\caption{An example of correction results at different levels}
\centering
\tabcolsep=0.37cm
\label{tab:1}
\begin{tabular}{cc}
\hline
Sentence Type              & Sentence Example                           \\ \hline
Original correct sentence  & \begin{CJK}{UTF8}{gbsn}他上周去了金字塔。\end{CJK}He went to the pyramid last week. \\
Incorrect sentence         & \begin{CJK}{UTF8}{gbsn}他上周取了金字。\end{CJK}He got golden character last week. \\
Character-level correction & \begin{CJK}{UTF8}{gbsn}他上周取了金子。\end{CJK}He got gold last week.             \\
Phrase-level correction    & \begin{CJK}{UTF8}{gbsn}他上周取了金字塔。\end{CJK}He got the pyramid last week.     \\ \hline
\end{tabular}
\end{table}

For example, the correction process from the semantic perspective is vulnerable for the subtle errors at character-level. For a pair of correct and incorrect sentences shown in Tab. \ref{tab:1}, the incorrect sentence contains a grammatical error (\begin{CJK}{UTF8}{gbsn}金字\end{CJK}/gold character of \begin{CJK}{UTF8}{gbsn}金字塔\end{CJK}/pyramid) at phrase-level and a spelling error (\begin{CJK}{UTF8}{gbsn}取\end{CJK}/take of \begin{CJK}{UTF8}{gbsn}去\end{CJK}/went with a similar pronunciation) at character-level. Moreover, the corrected results at character-level and phrase-level, in which both the corrected sentences are still incorrect in semantic logic, cannot be completely corrected even by further correction process. 

The example of the correction results demonstrates that the reliable detection process is significant for the correction method at phrase-level.
As the error detection methods used at character-level~\cite{zhang-etal-2020-spelling,zhang-etal-2021-correcting}, the semantics-based error detection network is also adequate for grammar error correction. Besides, the example also reflects the inconsistency of different processing level, the Chinese text correction process cannot be regarded as a simple combination of above two aspects of correction procedures because of the semantic polysemy of different parsing. Recently, the unified method of the above two aspects of strategies has developed. Malmi proposed the LaserTagger method by tagging characters with base tags and added phrases~\cite{malmi-etal-2019-encode}. Wherein, the base tag process is like error detection mentioned before, and the added phrase is analyzed based on the constructed semi-dynamic vocabulary at both character-level and phrase-level. LaserTagger tends to break the boundary between the character-level and phrase-level, even if the completely dynamic vocabulary list is closely related to the minimum k-union problem, which is NP-hard.

\subsection{Motivation and Contributions}

In this study, we proposed a novel correction strategy by taking the character-level and phrase-level correction as a unified semantic correction process, and in which, an adversarial multi-task learning method is introduced to effectively model the unified semantics of a sentence. Here, the masked character prediction and token scoring tasks are considered a couple of learning tasks in the multi-task learning method~\cite{caruana1997multitask}. Moreover, their generative adversarial learning~\cite{Goodfellow-Ian} relation, the masked language model as the generator, and the scoring language model as the discriminator, is also newly introduced to improve task robustness.

In addition, for the automatic poetry writing area~\cite{zhang-lapata-2014-chinese,ghazvininejad-etal-2016-generating,yi-etal-2018-chinese}, inspired by the polishing process~\cite{Yan-Rui,Deng_Wang_Liang_Chen_Xie_Zhuang_Wang_Xiao_2020}, we also introduce the polishing concept to realize the text correction based on the trained masked language model and scoring language model. Besides, because the sentence length of the common text is flexible which is different from poetry writing, the MCTS (Monte Carlo tree search~\cite{Kocsis-Levente,Browne-Cameron-B}) strategy is introduced to our method inspired by the idea in AlphaGo~\cite{silver2017mastering}.

\subsection{Achievements}
Thanks to the newly innovated adversarial multi-task learning strategy, the following contributions can be gained.

\begin{itemize}
\item
A new Chinese text correction framework with a unified correction process of character-level and phrase-level is proposed by means of a more robust sentence semantic modeling strategy.
\item
Two core models for text correction, masked language model and scoring language model, are designed as a group of adversarial multi-task learning models.
\item
Our experimental analysis indicates that the MCTS strategy should be valuable to improve the text correction quality for those sentences with multiple positions and character errors (or needing complex editing correction).
\end{itemize}

\section{Methodology}

\subsection{Task Formulation}
Given a sentence $\mathbb{S}=(c_1,\dots,\dot{c}_i,\dot{c}_{i+1},\dots,c_k)$ with some character errors $(\dot{c}_i,\dot{c}_{i+1})$ originated from the correct sentence $\mathrm{S}=(c_1,\dots,c_i,\dots,c_k)$. The correction task can be defined as a prediction task of generating a corrected sentence $\hat{\mathrm{S}}$ from the wrong sentence $\mathbb{S}$ with minimum semantic deviation.

\subsection{Adversarial Multi-task Learning}

\begin{figure}[t]
\centering
\includegraphics[width=\textwidth]{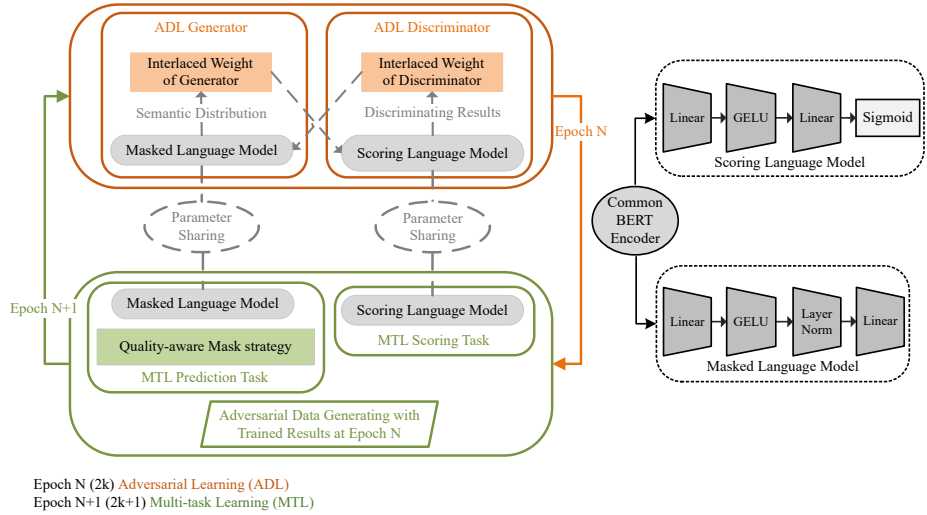}
\caption{The framework of adversarial multi-task learning method (Best view in color).} \label{fig1}
\end{figure}

\subsubsection{Structure of Adversarial Multi-task Learning}
We consider that adversarial multi-task learning is a model training strategy for two adversarial and cooperative model tasks. For Chinese text correction, two core task models, including the masked language model and the scoring language model, are considered in this study. Wherein, the masked language model is to predict reasonable characters in the masked position of a sentence, and the scoring language model is to examine the semantic quality of every character in its sentence context.

As shown in Fig. \ref{fig1}, a BERT~\cite{devlin-etal-2019-bert} encoder is introduced as the common encoder for scoring language model and masked language model. The adversarial multi-task learning is composed of the adversarial learning phase and multi-task learning phase. The multi-task learning aims to find the optimal semantic representation using the common encoder for different down-stream tasks. While, the adversarial learning is to enhance the robustness of classifier in distinguishing similar tokens in a sentence.

\subsubsection{The Multi-task Learning Phase}

Benefits from the structure of multi-task learning paradigm, the common encoder effectively prevent the model from the overfitting problem and the general semantic representation can enhance the modeling ability of the method. However, the general semantic representation from the encoder is vulnerable to adversarial attacks from the generative tokens which are semantic similar to original tokens~\cite{ruder2017overview}. Besides the introduction of the following adversarial learning phase, an additional adversarial data generation strategy is introduced in this phase.

\paragraph{Adversarial Data Generating Strategy}

\begin{figure}[t]
\centering
\includegraphics[width=\textwidth]{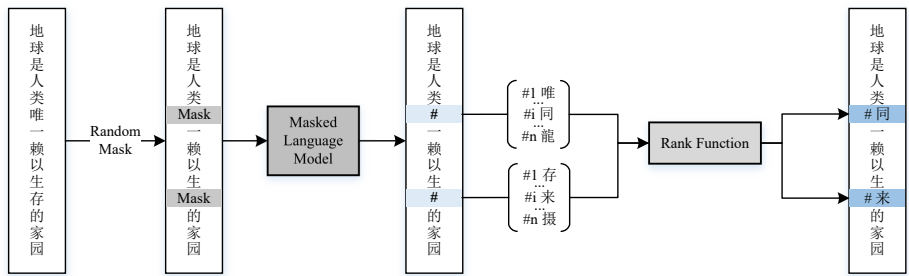}
\caption{Adversarial data generating strategy.}
\label{fig2}
\end{figure}

As shown in Fig. \ref{fig2}, the trained masked language model is firstly used to generate a group of ordered candidate token (character) sequence according to their confidence distributions on given masked positions in a sentence. Moreover, an index function is defined as following formula to select a token in a candidate token sequence
\begin{equation}
\label{eq:1}
Ik_t(\cdot)=e^{randI(\cdot)\cdot ln(Ct)}
\end{equation}
where, $Ik_t(\cdot)$ denotes an index value generation function, and $randI()$ is a uniform random function on $[0,1]$. $Ct$ is a constant, and the value 1000 is considered as default in this study. The generated tokens less than the threshold 20 are considered as positive token (has reasonable semantic), otherwise as negative token.

\subsubsection{The Adversarial Learning Phase}
In this part, the masked language model and the scoring language model are respectively considered as the generator and the discriminator in standard adversarial learning method. However, when the vanilla generative adversarial method is applied to natural language process problem, the excessive precision of the generator will lead to training confusion of the discriminator~\cite{Kevin-Clark-and-Thang}. Therefore, in order to keep the increasing balance with the training execution, the interlaced weights between the generator and the discriminator are introduced in our framework. The interlaced weights are used to dynamically adjust the learning strengths of different generated samples and discriminated samples according to the semantic similarity of the predicted tokens to correct tokens.

\paragraph{Interlaced Weights Definitions}
Here, we use $W_G$ to denote the interlaced weight matrix from the generator to the discriminator, and $W_D$ to denote the interlaced weight matrix from the discriminator to the generator. For $W_G$, it is used to adjust the training strength of the discriminator for different generative tokens, and lower values will be calculated for those generative tokens with higher semantic similarity to their correct original tokens. Concretely, the following equations can be defined.
\begin{equation}
\label{eq:2}
W_\mathcal{G}[i]=
\begin{cases}
\mathrm{sigmoid}(s_i)+0.5  & \text{ if } s_i \ge 0 \\
\mathrm{tanh}(s_i)  & \text{ if } s_i \le 0
\end{cases}
\end{equation}
\begin{equation}
\label{eq:3}
s_i=\frac{d_i [0]-d_i [Ik_t ]}{d_i [Ik_t ] }-S_g
\end{equation}
where, $s_i$ denotes the similarity score values of generative token from original correct token in the  $i$-th character position in a sentence, and  $d_i$ is corresponding confidence degree value vector related to the generative candidate token sequence by masked language model. Besides, $S_g$ is a hyperparameter with a default value 1.15 according to our experimental analysis.

Thus, for those generative tokens with high semantic similarity to their original correct tokens, they will be weakened in the discriminator training. That is, the discriminator will not be confused by those generative (adversarial) samples with high semantic reasonability but not originally correct.

For $W_D$, it is used to adjust the training strength of the generator, and the lower values will be determined that the discriminator cannot effectively distinguish the generated tokens and their original correct tokens.  In this study, the following equations can be defined.
\begin{equation}
\label{eq:4}
W_\mathcal{D} [i]=\frac{score_g [i]}{score_o [i]}
\end{equation}
For the $i$-th character in $S$, $score_g$ and $score_o$ denote the score vector of different positions respectively for the generative sentence and the original sentence predicted by the discriminator.

Therefore, $W_\mathcal{D}$ also reversely indicates the generative performance of the generator, and can subtly raise the training strength of those prediction tokens with high semantic similarity to original tokens but low discrimination scores.

\subsubsection{Training for Adversarial Multi-task Learning}
For the loss functions of adversarial multi-task learning, the objective of generative adversarial learning and the objective of multi-task learning should be respectively considered.
\paragraph{Objectives of Adversarial Learning}
According to the above discussions and standard entropy loss function, the following objectives are introduced.
\begin{equation}
\label{eq:5}
L_{\mathcal{G}}=L_{\mathcal{D} \sim \mathcal{G}}=\sum_{i \in R} W_{\mathcal{D}}[i]\{-x\left[\text { class }_{i}\right]+\log (\sum_{j=0}^{v s} e^{x[j]})\}
\end{equation}
\begin{equation}
\label{eq:6}
L_{\mathcal{D}}=L_{\mathcal{G} \sim \mathcal{D}}+L_{\mathcal{D} \sim \mathcal{D}}
\end{equation}
\begin{equation}
\label{eq:7}
L_{\mathcal{G} \sim \mathcal{D}}=\sum_{i=1}^{k} W_{\mathcal{G}}[i]\left[y_{i} \log \left(x_{i}\right)+\left(1-y_{i}\right) \log \left(1-x_{i}\right)\right]
\end{equation}
\begin{equation}
\label{eq:8}
L_{\mathcal{D} \sim \mathcal{D}}=\frac{\sum_{i \in C} \operatorname{sigmoid}\left(x_{i}\right)}{\sum_{j \in R} \operatorname{sigmoid}\left(x_{j}\right)}
\end{equation}
where, $k$ denotes the number of the total token (character) in a given sentence, and $R$ and $C$ are the index sets with generated tokens and original tokens. Besides, $L_{\mathcal{D}\sim \mathcal{G}}$ and $L_{\mathcal{G}\sim \mathcal{D}}$ reflect the interlaced loss objectives between the generator and discriminator. And, $L_{\mathcal{D}\sim \mathcal{D}}$ is the loss function of discriminating the generative tokens from original correct tokens. For these objectives, all parameters except in the common encoder should be optimized.

\paragraph{Objectives of Multi-task Learning}
The general objectives are introduced as follows
\begin{equation}
\label{eq:9}
L_{M T L}=L_{P S}+L_{M L}
\end{equation}
\begin{equation}
\label{eq:10}
L_{S}=\sum_{i=1}^{k} y_{i} \log \left(x_{i}\right)+\left(1-y_{i}\right) \log \left(1-x_{i}\right)
\end{equation}
\begin{equation}
\label{eq:11}
L_{M L}=\sum_{i=1}^{k}-x[\operatorname{class_{i}]}+\log (\sum_{j=0}^{vs} e^{x[j]}).
\end{equation}
where $L_{PS}$ and $L_{ML}$ are the objectives of the masked language model and the scoring language model respectively. Similarly, $k$ denotes the total token (character) numbers in a given sentence, $vs$ is the size of the vocabulary list.

\subsection{Chinese Text Correction with Semantic Error Detection}

In order to realize the correction of a sentence with possible errors, the Monte Carlo tree search (MCTS) strategy is introduced to find and correct possible error characters. Wherein, all possible positions with error characters can be found by means of the trained scoring language model, and the most reasonable (correct) sentence can be recommended based on the combined use of the trained masked language and scoring language models. In summary, the following Algorithm 1 can be proposed.

In above Algo. \ref{algo:1}, the two-fold scoring and prediction searching\footnote{In Algo. \ref{algo:1}, the default values of width (step 2) and depth (step 3) are set as 2 and 4.} has to be executed, so the computing efficiency will not satisfy the practical requirements. Furthermore, a policy network inspired by the idea of AlphaGo is introduced to speed up the error position searching computation. In our policy network, the most possible error range in a sentence is defined as the network output.

\begin{algorithm}[t] 
  \begin{algorithmic}[1] 
  \renewcommand{\algorithmicrequire}{ \textbf{Input}: Sentence $\mathbb{S}={c_1,c_2,...,\dot{c}_i,\dot{c}_{i+1},...,c_k }$ with errors ${\dot{c}_i,\dot{c}_{i+1}}$}
  \REQUIRE
  \renewcommand{\algorithmicrequire}{ \textbf{Output}: Corrected sentence $\hat{\mathrm{S}}_{\mathrm{MCST}}={c_1,...,c_i,,...,c_{k-1} }$}
  \REQUIRE
  \STATE {Perform the \textbf{Scoring Language Model} on $\mathbb{S}$ to get the $score_\mathbb{S}={s_1,\dots,s_{k+1}}$ and the character position with the maximum score $p_{m}$}
  \FOR{$p_s\leftarrow p_{m}$ and $p_e\leftarrow p_{m}$ \textbf{to} $p_{m}-width$ and $p_{m}+width$}
  \FOR{$num\leftarrow 0$ \textbf{to} $depth$}
  \STATE{
  Replace $\{p_s,\dots,p_e\}$ with $\{\overbrace{[\mathrm{MASK}],\dots,[\mathrm{MASK}]}^{num*[\mathrm{MASK}]}\}$ to get $\mathbb{S}_{p_s,p_e}$
  }
  \STATE{
  Perform the \textbf{Masked Language Model} on $\mathbb{S}_{p_s,p_e}$ to get $\hat{\mathrm{S}}_{p_s,p_e}=c_1,...,\hat{c}_i,\hat{c}_{i+num-1},...,c_{k-1}$
  }
  \STATE{
  Perform the \textbf{Scoring Language Model} on $\hat{\mathrm{S}}_{p_s,p_e}$, $score_{\hat{\mathrm{S}}}=\mathrm{SLM}(\hat{\mathrm{S}}_{p_s,p_e})/k$
  }
  \STATE{
  Append $score_{\hat{\mathrm{S}}}$ to $list_{\hat{\mathrm{S}}}$
  }
  \ENDFOR
  \ENDFOR
  \STATE {
  Return $\hat{\mathrm{S}}_{\mathrm{MCST}}|score_{\hat{\mathrm{S}}}=\mathrm{min}(list_{\hat{\mathrm{S}}})$
  }
   
  \end{algorithmic}
  \caption{Basic Correction Process with Adversarial Multi-task Learning}
  \label{algo:1}
\end{algorithm}

In our policy network, the network output is considered to be the most possible error range for a wrong sentence. Here, all wrong sentences are generated by randomly replacing multiple character strings (with a maximum length of 4) in original correct sentences, in which the lengths of the original and replaced character strings may be different. Moreover, the corrected sentences of those wrong sentences can be predicted according to Algo. \ref{algo:1}. Thus, for a generated wrong sentence, the most possible error character can be recognized based on the trained scoring language model. Then we may introduce $(Is_{w\sim o},Ie_{w\sim o})$ to denote the start and end position index values of the error string containing the most possible error character in the wrong sentence. Similarly, $(Is_{w\sim cw},Ie_{w\sim cw})$ is used to indicate the start and end position index values of the corresponding string in the wrong sentence of the corrected string in the corrected sentence by Algo. \ref{algo:1}, in which the corresponding string still contains that most possible error character.

Moreover, for a wrong sentence and its most error character string containing the most possible error character evaluated by the trained scoring language model, the following computations can be defined as
\begin{equation}
\label{eq:12}
\left[S_{l}, S_{h}\right]=\left[\min \left(I s_{w \sim o}, I s_{w-c w}\right), \max \left(I s_{w \sim o}, I s_{w \sim c w}\right)\right]
\end{equation}
\begin{equation}
\label{eq:13}
\left[E_{l}, E_{h}\right]=\left[\min \left(I e_{w \sim o}, I s_{w-c w}\right), \max \left(I e_{w \sim o}, I e_{w \sim c w}\right)\right]
\end{equation}
here, $[S_l,S_h ]$ and $[E_l,E_h ]$ as possible position range indicators of an error character string are introduced to make the policy network can output a reasonable error position range for a detected sentence.

In addition, an extra loss coefficient is introduced to indicate the performance on error recognition coverage and corresponding correction accuracy of algorithm 1, which is calculated by the following formula

\begin{equation}
\label{eq:14}
\mu=\frac{\operatorname{len}(\operatorname{char}_{I S_{W \sim C W}-I e_{W \sim C W}}{ }{\cap \operatorname{char}}{ }_{I s_{W \sim O}-I e_{W \sim O}})^{2}}{\operatorname{len}\left(\operatorname{char}_{Is_{W \sim C W}-Ie_{W \sim C W}}\right) \operatorname{len}\left(\operatorname{char}_{I S_{W \sim O}-I e_{W \sim O}}\right)}
\end{equation}
Where, $char_{Is_{w\sim cw}-Ie_{w\sim cw}}$ and $char_{Is_{w\sim o}-Ie_{w\sim o}}$ are strings in range of ${Is_{w\sim cw}-Ie_{w\sim cw}}$ and ${Is_{w\sim o}-Ie_{w\sim o}}$, $len(\cdot)$ function is used to calculate the character-length of the string. Based on the loss coefficient $\mu$ and the position ranges of start and end labels, we can jointly learn the start position prediction task and end position prediction task to train the policy network by minimizing the following objectives.
\begin{equation}
\label{eq:15}
L=0.5 L_{R_{S}}+0.5 L_{R_{e}}
\end{equation}
\begin{equation}
\label{eq:16}
L_{R_{s}}=\mu\left[\operatorname{sam}\left(x_{s}\right)-S_{l}\right]^{2} e^{S_{l}-\operatorname{sam}\left(x_{s}\right)}+(1-\mu)\left[\operatorname{sam}\left(x_{s}\right)-S_{h}\right]^{2} e^{\operatorname{sam}\left(x_{s}\right)-S_{h}}
\end{equation}
\begin{equation}
\label{eq:17}
L_{R_{e}}=\mu\left[\operatorname{sam}\left(x_{e}\right)-E_{l}\right]^{2} e^{E_{l}-\operatorname{sam}\left(x_{e}\right)}+(1-\mu)\left[\operatorname{sam}\left(x_{e}\right)-E_{h}\right]^{2} e^{\operatorname{sam}\left(x_{e}\right)-E_{h}}
\end{equation}
Where $x_{start}$ and $x_{end}$ are the prediction results of the policy network, and $\operatorname{sam}(\cdot)$ function is soft-argmax function~\cite{nibali2018numerical}. The policy network faster in prediction compared to the standard MCTS correction process. Additionally, the policy network is semantic sensitive, which results in a more flexible usage scenario.

\section{Experimental Result}
\subsection{Dataset}
The datasets in our experiments are CLUE (Chinese Language Understanding Evaluation Benchmark~\cite{xu-etal-2020-clue}), CGED-2018 (Chinese Grammatical Error Diagnosis~\cite{rao-etal-2018-overview}), and our Xuexi dataset.

The corpora in CGED-2018 were taken from the testing result of HSK (Pinyin of Hanyu Shuiping Kaoshi, Test of Chinese Level). The dataset contains different types of grammar errors, which can ideally reflect the situation of the writing errors faced in our daily. CLUE is the largest language understanding corpus in Chinese, in which the Chinese Wikipedia dataset is selected and is mixed with multiple languages characters representing a complex semantic environment. Xuexi dataset are collected by ours from the most prominent Chinese political news website "\begin{CJK}{UTF8}{gbsn}学习强国\end{CJK}/Xuexi Qiangguo"\footnote{\url{https://www.xuexi.cn/}}, which is managed by the Central Propaganda Department of the Communist Party of China. Xuexi dataset can reflect the usage scenario for people in daily working.

For model training, 200,000 sentences are randomly selected from Xuexi and CLUE datasets. In the main evaluation experiment, we randomly select 1,000 samples from each dataset mentioned above. In addition, for those 1,000 selected sentences from Xuexi and CLUE corpus, corresponding wrong sentences are constructed by using a similar strategy of generating wrong sentences for policy network training.

\subsection{Training Settings}
Our experimental environment is built on NVIDIA Tesla V100 (16GB GPU RAM), using a Transformer Package based on PyTorch. The pretrained BERT-base model for Chinese is employed to initialize the language models. The model is trained using AdamW~\cite{loshchilov2018decoupled}. We set the initial learning rate to 0.00002 and set the first 10k adjustments to the warm-up stage, and then linearly adjusted the learning rate. For the application scenarios of the text correction, the parameters in the original BERT-Base model are partly customized. We set the length to 64 Chinese and English characters, keeping the remaining hyperparameters consistent with BERT-base. In the training process of the policy network, we set the dropout rate to 0.5 to train the generalization ability of the model.

\subsection{Ablation Results}

\begin{table}[t]
\caption{Ablation Results}
\label{tab:2}
\centering
\tabcolsep=0.28cm
\setlength{\textwidth}{1.5mm}{
\begin{tabular}{ccccccc}
\hline
\multirow{2}{*}{Dataset} & \multirow{2}{*}{Method}         & \multicolumn{4}{c}{MLM}    & SLM  \\ \cline{3-7} 
                         &                                 & Acc. & Prec. & Rec. & F1.  & Acc. \\ \hline
\multirow{4}{*}{CLUE}    & Normal Supervised-Learning      & 58.5 & 42.2  & 42.0 & 40.3 & 69.8 \\
                         & Multi-task Learning             & 61.3 & 42.7  & 44.7 & \textbf{44.0} & 74.1 \\
                         & Generative Adversarial Learning & 52.9 & 34.5  & 37.4 & 35.4 & 73.6 \\
                         & Adversarial Multi-task Learning & \textbf{62.9} & \textbf{45.0}  & \textbf{44.9} & 43.3 & \textbf{75.7} \\ \hline
\multirow{4}{*}{Xuexi}   & Normal Supervised-Learning      & 49.2 & 33.1  & 33.2 & 31.8 & 71.9 \\
                         & Multi-task Learning             & 59.3 & 42.7  & 41.6 & 40.3 & 72.9 \\
                         & Generative Adversarial Learning & 51.4 & 33.1  & 32.7 & 31.3 & 64.9 \\
                         & Adversarial Multi-task Learning & \textbf{61.1} & \textbf{44.3}  & \textbf{43.1} & \textbf{41.9} & \textbf{74.8} \\ \hline
\end{tabular}}
\end{table}

We set up an ablation experiment on CLUE Dataset and Xuexi Dataset to evaluate the performance of the adversarial multi-task learning method. For evaluating the scoring language model, we sort the score sequence of the sentence and extract the first K positions with high rank, then the scoring accuracy is calculated by comparing with the error positions in the sentence.

As shown in Tab. \ref{tab:2}, the result of the adversarial multi-task learning method has a significant improvement compared with other models, which indicates that our adversarial multi-task learning method is efficient in modeling ability improvement.

\subsection{Main Results}

Here, several open-source Chinese detection and correction tools are introduced to examine the correction performance of different methods. The Founder detection tool as a classical technique in Chinese proofreading is performed to detect the number of errors in corrected sentences. The number of errors reflects the error correction capability of the methods. The NLP tools from Baidu's artificial intelligence platform are also used, in which the language perplexity (PPL) computation can evaluate the fluency of sentences based on every word in the sentences, and the DNN score can evaluate the possibility of every word in the sentence. Wherein, the normalized performance values in illustrated results are calculated by the ratio with respect to PPL and DNN score values of original sentences. BLEU is a classical evaluation matrix, which is widely used in the evaluation of translation task. It reflects the ability of restoring an error sentence to its original sentence for text correction methods.

In our correction performance analysis, the hard mode named as VarLen scenario~\cite{li-shi-2021-tail} is also considered, in which the two length values of a wrong sentence and its ideal correct sentence are different. However, some of the existing text correction methods can only perform the correction for the FixLen scenario. So, the following methods are adopted as comparison including ELECTRA~\cite{Kevin-Clark-and-Thang}, ERNIE~\cite{zhang-etal-2019-ernie}, MacBERT~\cite{cui-etal-2020-revisiting}, and TtT~\cite{li-shi-2021-tail} as a state-of-the-art method in variable-length correction as the baseline.

\begin{table}[t]
\caption{Main Results}
\label{tab:3}
\centering
\tabcolsep=0.3cm
\begin{tabular}{cccccl}
\hline
Dataset                & Method     & Founder Detection & Baidu PPL      & Baidu DNN      & \multicolumn{1}{c}{BLEU} \\ \hline
\multirow{5}{*}{Xuexi} & MacBERT    & 78                & 0.320          & 0.638          & 0.703                    \\
                       & ERNIE      & 105               & 0.262          & 0.560          & 0.687                    \\
                       & ELECTRA    & 103               & 0.257          & 0.561          & 0.694                    \\
                       & TtT        & 122               & 0.252          & 0.557          & 0.694                    \\
                       & Our Method & \textbf{7}        & \textbf{0.813} & \textbf{0.929} & \textbf{0.738}           \\ \hline
\multirow{5}{*}{CLUE}  & MacBERT    & 61                & 0.498          & 0.707          & \textbf{0.685}           \\
                       & ERNIE      & 119               & 0.559          & 0.694          & 0.663                    \\
                       & ELECTRA    & 84                & 0.507          & 0.714          & 0.678                    \\
                       & TtT        & 98                & 0.528          & 0.717          & 0.679                    \\
                       & Our Method & \textbf{6}        & \textbf{0.814} & \textbf{0.958} & 0.629                    \\ \hline
\multirow{5}{*}{CGED}  & MacBERT    & 22                & \textbf{0.846} & 0.920          & 0.720                    \\
                       & ERNIE      & 41                & 0.795          & 0.921          & 0.712                    \\
                       & ELECTRA    & 36                & 0.774          & 0.916          & \textbf{0.819}           \\
                       & TtT        & 44                & 0.766          & 0.914          & 0.818                    \\
                       & Our Method & \textbf{5}        & 0.766          & \textbf{1.021} & 0.643                    \\ \hline
\end{tabular}
\end{table}

As shown in Tab. \ref{tab:3}, our method gained better performance compared to the baseline in various evaluation indexes. Even so, because the structure and length of the wrong sentences are not much different from the original sentences, our method did not get a significant improvement in BLEU (for restoring ability evaluation). But in Founder detection performance, our method has an overwhelming advantage over all other methods, which represents that our method can effectively discover the semantic errors in the sentences and correct them under appropriate semantics.

\begin{table}[t]
\caption{Examples of the Correction Methods}
\label{tab:4}
\centering
\renewcommand{\arraystretch}{1.2} 
\tabcolsep=0.34cm
\begin{tabular}{ccc}
\hline
Num.               & Method                      & e.g                                                             \\ \hline
\multirow{2}{*}{1} & \multirow{2}{*}{Original}   & \begin{CJK}{UTF8}{gbsn}试着在国际\uline{竞争中}\uwave{掌握}更大话语权。\end{CJK}                                                \\
                   &                             & Trying to \uwave{control} a greater power \uline{in} international \uline{competition}. \\ \hdashline
\multirow{2}{*}{2} & \multirow{2}{*}{Wrong}      & \begin{CJK}{UTF8}{gbsn}试着在国际\uline{竞争力}\uwave{具己}更大话语权。\end{CJK}                                                \\
                   &                             & Trying to \uwave{'juji'} a greater power international \uline{competitiveness}. \\ \hdashline
\multirow{2}{*}{3} & \multirow{2}{*}{MacBERT}    & \begin{CJK}{UTF8}{gbsn}试着在国际\uline{竞争力}\uwave{具有}更大话语权。\end{CJK}                                                \\
                   &                             & Trying to \uwave{have} a great status international \uline{competitiveness}.    \\ \hdashline
\multirow{2}{*}{4} & \multirow{2}{*}{ERNIE}      & \begin{CJK}{UTF8}{gbsn}试着在国际\uline{竞争力}\uwave{具己}更大话语权。\end{CJK}                                                \\
                   &                             & Trying to \uwave{'juji'} a greater power international \uline{competitiveness}. \\ \hdashline
\multirow{2}{*}{5} & \multirow{2}{*}{ELECTRA}    & \begin{CJK}{UTF8}{gbsn}试着在国际\uline{竞争力}\uwave{具己}更大话语权。\end{CJK}                                                \\
                   &                             & Trying to \uwave{'juji'} a greater power international \uline{competitiveness}. \\ \hdashline
\multirow{2}{*}{6} & \multirow{2}{*}{TtT}        & \begin{CJK}{UTF8}{gbsn}试着在国际\uline{竞争力}\uwave{具己}更大话语权。\end{CJK}                                                \\
                   &                             & Trying to \uwave{'juji'} a greater power international \uline{competitiveness}. \\ \hdashline
\multirow{2}{*}{7} & \multirow{2}{*}{Our Method} & \begin{CJK}{UTF8}{gbsn}试着在国际\uline{竞争力中}\uwave{争得}更大话语权。\end{CJK}                                               \\
                   &                             & \uwave{Striving for} a greater voice \uline{in} international \uline{competitiveness}.  \\ \hline
\end{tabular}
\end{table}

In Tab. \ref{tab:4}, an example is presented including an original correct sentence, a typical wrong sentence generated in our experiment, and five corrected results by four comparable correction methods and our method. For the above different corrected results, the corrected result of our method might be the most reasonable in sentence semantics. Concretely, for the character string with wave underlines in sentences, our method accurately understands the emotion of the given wrong sentence and finds similar words (\begin{CJK}{UTF8}{gbsn}争得\end{CJK}/strive for). Similarly, for the character string with straight under-lines, only our method can give a correction solution that the (\begin{CJK}{UTF8}{gbsn}中\end{CJK}/in) is added. Of course, the last result of our method still has not completely repaired the wrong sentence to the ideal result (original correct sentence), which also reflects the complexity and hardness of the Chinese text correction task.

\section{Conclusion}
For Chinese text correction task with complex semantic errors, we proposed a novel correction method by introducing an adversarial multi-task learning strategy. The proposed method can model Chinese sentences into a unified semantic feature and effectively resolve the modeling confusion of similar characters in semantics. In addition, a policy network is introduced to gain the high searching quality and efficiency of error positions. The experimental results on three datasets clearly indicate the significant advantages of our method in the VarLen scenario.

In the future, we plan to enhance the explainable modeling ability for Chinese text correction by developing explainable scoring language models.

\subsubsection{Acknowledgements}
This work was supported in part by the grants from National Natural Science Foundation of China (Grant no. 61872166) and Six Talent Peaks Project of Jiangsu Province of China (2019 XYDXX-161).

\bibliographystyle{splncs04}
\bibliography{mybibliography}
\end{document}